\documentclass[a4paper, 10 pt, conference]{ieeeconf} 
\IEEEoverridecommandlockouts                             
\usepackage{color}
\usepackage{graphicx} 
\usepackage{mathtools} 
\usepackage{siunitx}
\usepackage[backend=biber, style=ieee,maxbibnames=99]{biblatex}
\AtBeginBibliography{\footnotesize}
\title{\LARGE \bf Hybrid Soft-Rigid Continuum Robot Inspired by Spider Monkey Tail}

\author{Mary C. Doerfler$^{1}$, Katalin Schäffer$^{1,2}$, Margaret M. Coad$^{1}$
\thanks{$^{1}$Department of Aerospace and Mechanical Engineering, University of Notre Dame, Notre Dame, IN 46556, USA {\tt\small \{mdoerfle, kschaff2, mcoad\}@nd.edu}}%
\thanks{$^{2}$Pázmány Péter Catholic University, Budapest, Hungary}%
}

\begin{document}

\maketitle
\thispagestyle{empty}
\pagestyle{empty}

\begin{abstract}

Spider monkeys (genus \textit{Ateles}) have a prehensile tail that functions as a flexible, multipurpose fifth limb, enabling them to navigate complex terrains, grasp objects of various sizes, and swing between supports. Inspired by the spider monkey tail, we present a life size hybrid soft-rigid continuum robot designed to imitate the function of the tail. Our planar design has a rigid skeleton with soft elements at its joints that achieve decreasing stiffness along its length. Five manually-operated wires along this central structure control the motion of the tail to form a variety of possible shapes in the 2D plane. Our design also includes a skin-like silicone and fabric tail pad that moves with the tail's tip and assists with object grasping. We quantify the force required to pull various objects out of the robot's grasp and demonstrate that this force increases with the object diameter and the number of edges in a polygonal object. We demonstrate the robot's ability to grasp, move, and release objects of various diameters, as well as to navigate around obstacles, and to retrieve an object after passing under a low passageway. 

\end{abstract}

\section{Introduction}

Among New World primates, the spider monkey has the most advanced prehensile tail, heavily used in what Rosenberger et al. describe as “acrobatic locomotion” \cite{Rosenberger2008}, making them an appropriate choice for a source of bio-inspiration. Spider monkeys perform several functions with their tails, including brachiation (swinging between tree branches using arms and tail), providing balance, extending reach, suspension, and manipulating objects. The dexterity and strength of the tail are desirable characteristics in a robotic manipulator. A robot imitating the function of a spider monkey tail might prove useful in many circumstances: for instance, a robotic tail might integrate with another system to provide balance while navigating unstable terrain inaccessible to humans (e.g., tree canopies or collapsed buildings), enabling a robot to swing over chasms and grasp various supports during its motion. A prehensile-tailed robot might access terrains inaccessible to a wheeled or legged robot.

\begin{figure}[tb]
      \centering
      \includegraphics[width=\columnwidth]{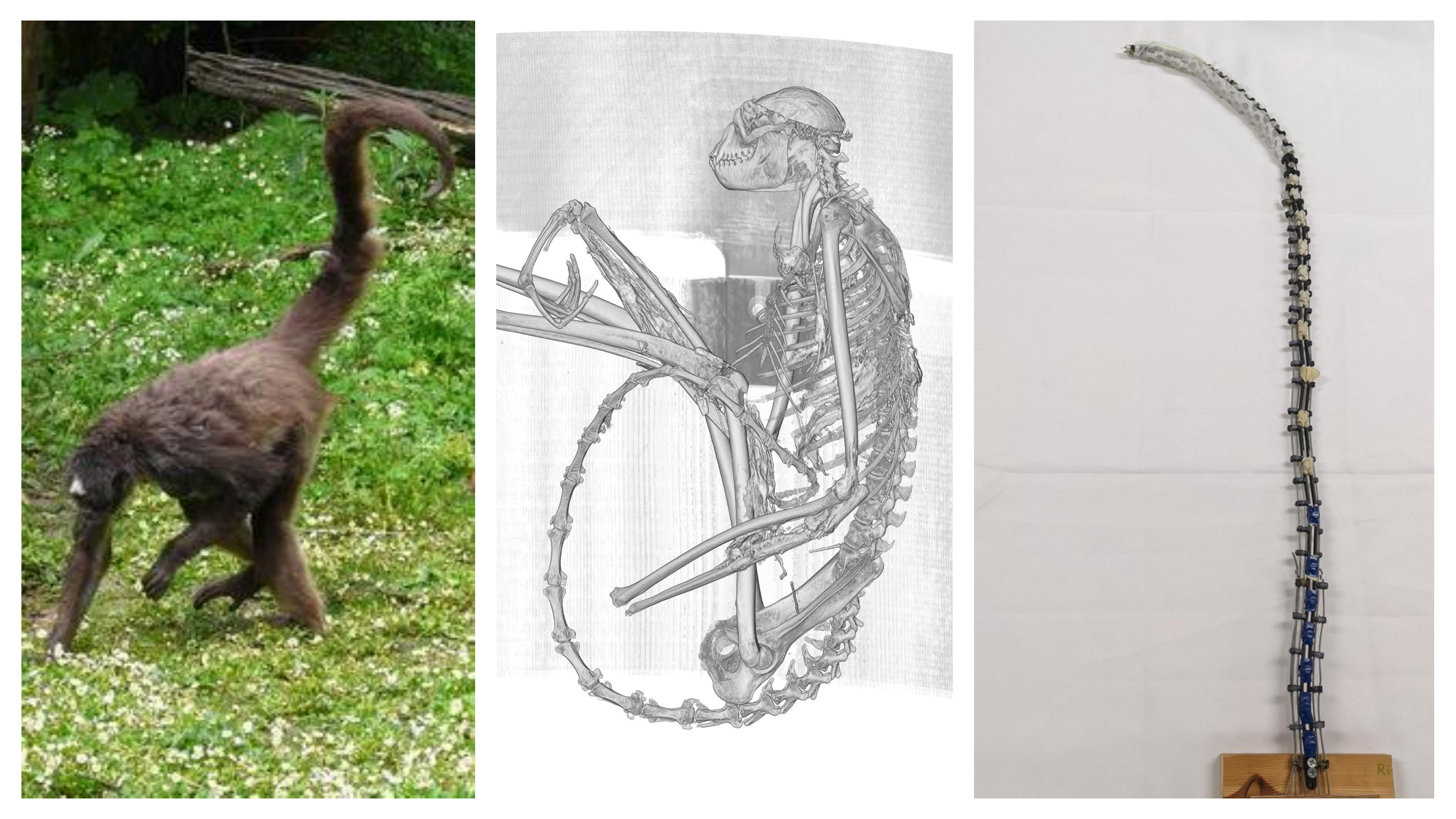}
     
      \caption{The spider monkey and our spider monkey tail-inspired robot. (\textit{left)} Brown spider monkey (\textit{Ateles hybridus}) using its tail for balance during clambering \cite{FinalMonkeyPic}. (\textit{middle}) CT scan of white-bellied spider monkey (\textit{Ateles belzebuth}), showing its skeletal structure \cite{CTScan}. (\textit{right}) Our life size spider monkey tail-inspired robot.
      }
      \label{GlamorShot}
      \vspace{-0.5cm}
  \end{figure}

Use of soft robotic technology has the potential to make robots lightweight, compliant, and cost-effective, while rigid components can provide structural stability. In this paper, we introduce a compact hybrid soft-rigid robot design which mimics key structural features of the biological monkey tail with the goal of also imitating its functionality for the described applications. Figure~\ref{GlamorShot} shows a biological spider monkey, a CT scan of a monkey's body, and our bio-inspired tail robot. Some biological tail features were implemented differently or left overseen by previous bio-inspired continuum robot designs, so we begin by providing a comparison of these robots with our own. Then, we describe the structure of our bio-inspired robotic tail, including the vertebral column, the soft elements providing varying degrees of stiffness, the actuation mechanism, and a novel tail pad.
Finally, we quantify the grasping force and demonstrate the dexterity of the tail while performing object grasping and retrieval tasks.

\section{Background}
A range of robots have been developed to imitate prehensile or flexible limbs with grasping capabilities. Arachchige and Godage~\cite{Arachchige2022} recently introduced a design for a spider monkey tail-inspired robot using McKibben-type pneumatic artificial muscles and a rigid-linked backbone. This design and four other bio-inspired continuum robot designs are compared in Table 1: an ostrich neck-inspired design \cite{Mochiyama2022}, a woodpecker tongue-inspired design \cite{Matsuda2022}, a seahorse tail-inspired design \cite{Holt2017}, and an elephant trunk-inspired design \cite{Hannan2000}. Each design incorporates elements of structural stiffness or rigidity along with flexible continuum elements. 

\begin{table*}
\caption{Bio-Inspired Continuum Robots with Soft and Rigid Elements and Grasping Capability}
\label{specs}
\centering
\begin{tabular}{|>{\centering\arraybackslash}p{2cm}|>{\centering\arraybackslash}p{2cm}|>{\centering\arraybackslash}p{2cm}|>{\centering\arraybackslash}p{2cm}|>{\centering\arraybackslash}p{2cm}|>{\centering\arraybackslash}p{2cm}|}
\hline
\textbf{Author, Year} & \textbf{Bio-Inspiration}& \textbf{Scale} & \textbf{Structure} & \textbf{Method of Actuation} & \textbf{Object Grasping Method}\\
\hline
\hline
Our robot & Spider monkey tail & Life size & Rigid links with soft material around rotational joints & Hand-operated wires & Whole-arm grasping with soft tail pad \\
\hline
Arachchige and Godage, 2022 \cite{Arachchige2022} & Spider monkey tail & Partial tail section, approx. life size & Commercial cable guide dress pack & McKibben pneumatic artificial muscles & Optional gripper attached to end \\
\hline
Mochiyama et al., 2022 \cite{Mochiyama2022} & Ostrich neck  & Life size & Rigid links with free rotational joints & Hand-operated wires & Head with beak attached to end of neck\\
\hline
Matsuda et al., 2022 \cite{Matsuda2022} & Woodpecker tongue & Larger than life & Flexible rack gear & Driving units running along rack gears & Whole-arm grasping \\
\hline
Holt, 2017 \cite{Holt2017} & Seahorse tail & Partial tail section, larger than life & 3D printed square ring segments & McKibben pneumatic artificial muscles & Whole-arm grasping if length of tail extended \\
\hline
Walker and Hannan, 2000 \cite{Hannan2000} & Elephant Trunk & Smaller than life & Segments linked with 2-DOF joints; springs for stiffness & Motor-operated wires & Whole-arm grasping \\
\hline

\end{tabular}\\
\vspace{5 pt}
\end{table*}

Our life size design imitates a biological spider monkey tail. Its structure employs rigid links with soft material around its rotational joints to provide structural stiffness. We use wires to manually actuate our robot and perform whole-arm grasping with the aid of a soft tail pad. In contrast, Arachchige and Godage's spider monkey tail-inspired design~\cite{Arachchige2022} was made on a limited scale, used a different actuation method than our design, and did not perform whole-arm grasping.  Mochiyama et al.'s ostrich neck-inspired design~\cite{Mochiyama2022} used similar actuation methods to our robot but did not perform whole-arm grasping. Matsuda et al.'s woodpecker tongue-inspired robot~\cite{Matsuda2022} used rack gears and driving units, a different actuation method than our robot. Holt's seahorse tail robot~\cite{Holt2017} imitated the morphological structure of a seahorse's tail on a limited scale, without grasping demonstrations. Walker and Hannan's elephant trunk robot~\cite{Hannan2000} used springs to provide structural stiffness; our design uses soft materials that are more lightweight and compact than springs. As compared to the other continuum robots evaluated, our robot differs in its method of creating structural stiffness and in its use of whole-arm grasping with a soft tail pad. 

\section{Spider Monkey Tail Anatomy and Use}
As inspiration for our robot design, we describe here the anatomy of the biological spider monkey tail, as well as the ways these monkeys use their tails in their environment.

\subsection{Anatomy}
The spider monkey tail reaches a length of about a meter, with typically 33 vertebrae along its length \cite{Chang1947}. The vertebrae diminish in size along the length of the tail. Chang and Ruch \cite{Chang1947} conceptualize the spider monkey tail in three segments: a proximal (base) region, a middle region, and a distal (tip) region. The proximal region includes vertebrae 1-8 and is relatively rigid, with well-developed processes (protrusions where tendons attach) on its vertebrae. The middle region includes vertebrae 9-17 and is relatively flexible, with longer and slenderer vertebrae as compared to the proximal region. The distal region, from vertebra 18 to the end of the tail, is the most flexible, mobile region. The vertebrae taper in size to the end, with the smallest bones about the size of a grain of barley. Muscles are radially distributed around the tail. The distal region also includes a skin-covered tail pad on the ventral surface; the tail pad has a texture like the palm of a human hand and assists in gripping supports and objects.

\subsection{Uses of Tail in Environment}
\begin{figure}[tb]
      \centering
      \includegraphics[width=0.9\columnwidth]{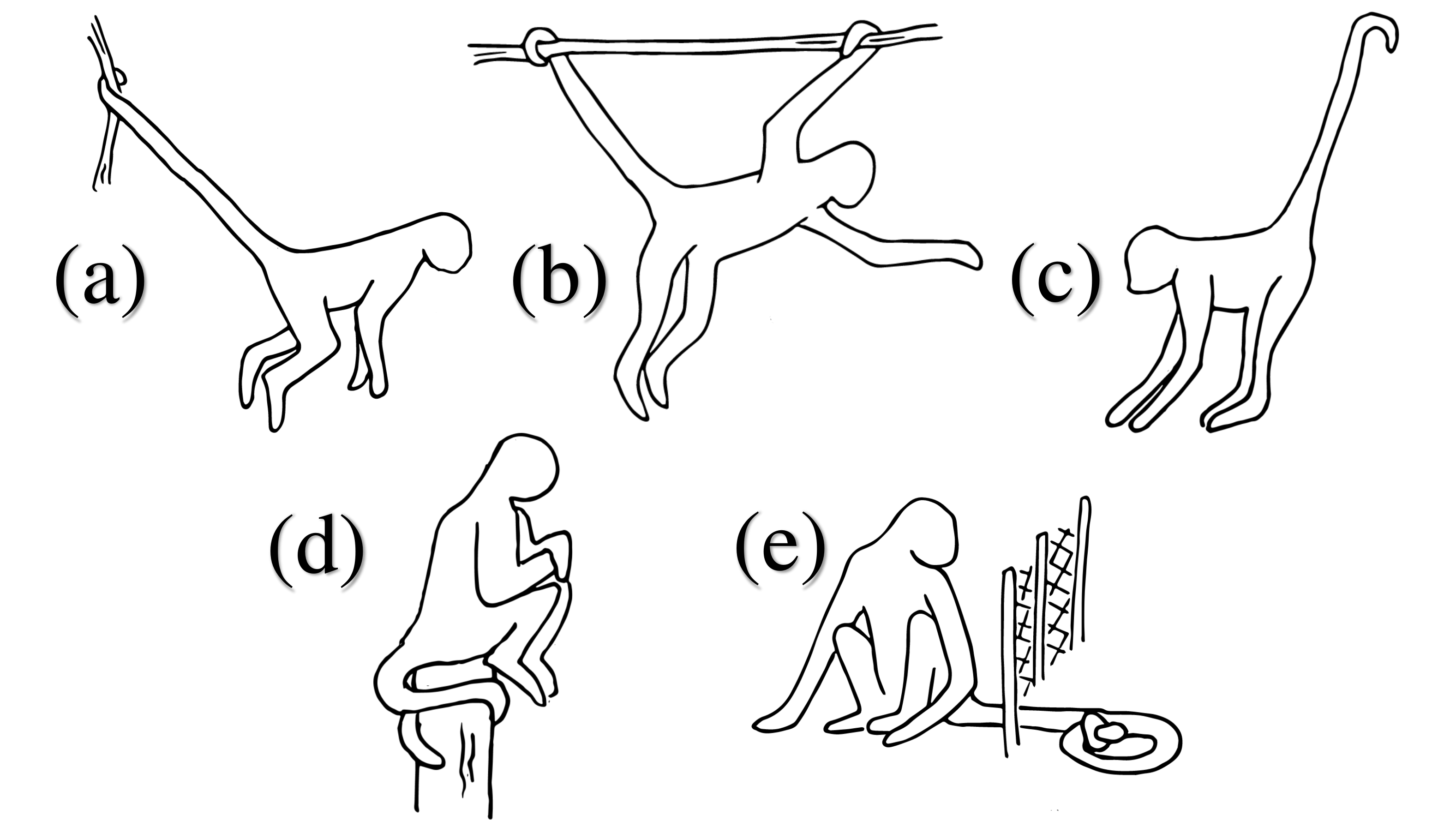}
     
      \caption{Uses of tail in biological spider monkeys. (a) Static support while swinging from tail alone. (b) Dynamic support while climbing using limbs and tail. (c) Balance assistance while clambering with tail upright. (d) Static support while resting. (e) Object retrieval for objects only reachable by tail.
      }
      \label{TailUse}
      \vspace{-0.5cm}
  \end{figure}
  
Spider monkeys employ their tails in a wide range of movements and behaviors (Figure~\ref{TailUse}). Their natural habitats include forests in Central and South America, where they spend the majority of their time in upper canopies and emergent strata (i.e., the tallest layers of the forest). Youlatos \cite{Youlatos2008} identifies common tail uses as swinging or suspension from the tail and/or limbs, brachiation (swinging dynamically between supports using limbs and tail), clambering (horizontal climbing across supports of various orientations), and bridging (moving between tree peripheries). Nelson and Kendall \cite{Nelson2018} also note goal-oriented tasks such as retrieving high-value morsels of food from locations only accessible by tail, perhaps due to a low passageway that permits a tail but not a hand. We observed and documented many of these behaviors at the Potawatomi Zoo in South Bend, IN, USA (see the associated video for this paper). The monkeys in residence (\textit{Ateles geoffroyi vellerasus}) use their tails nearly constantly, whether to provide assistance during clambering, to fully suspend the weight of their bodies while swinging between branches, or to wrap around the environment to support a resting posture.

\section{Mechanical Design}
\begin{figure}[tb]
      \centering
      \includegraphics[width=.85\columnwidth]{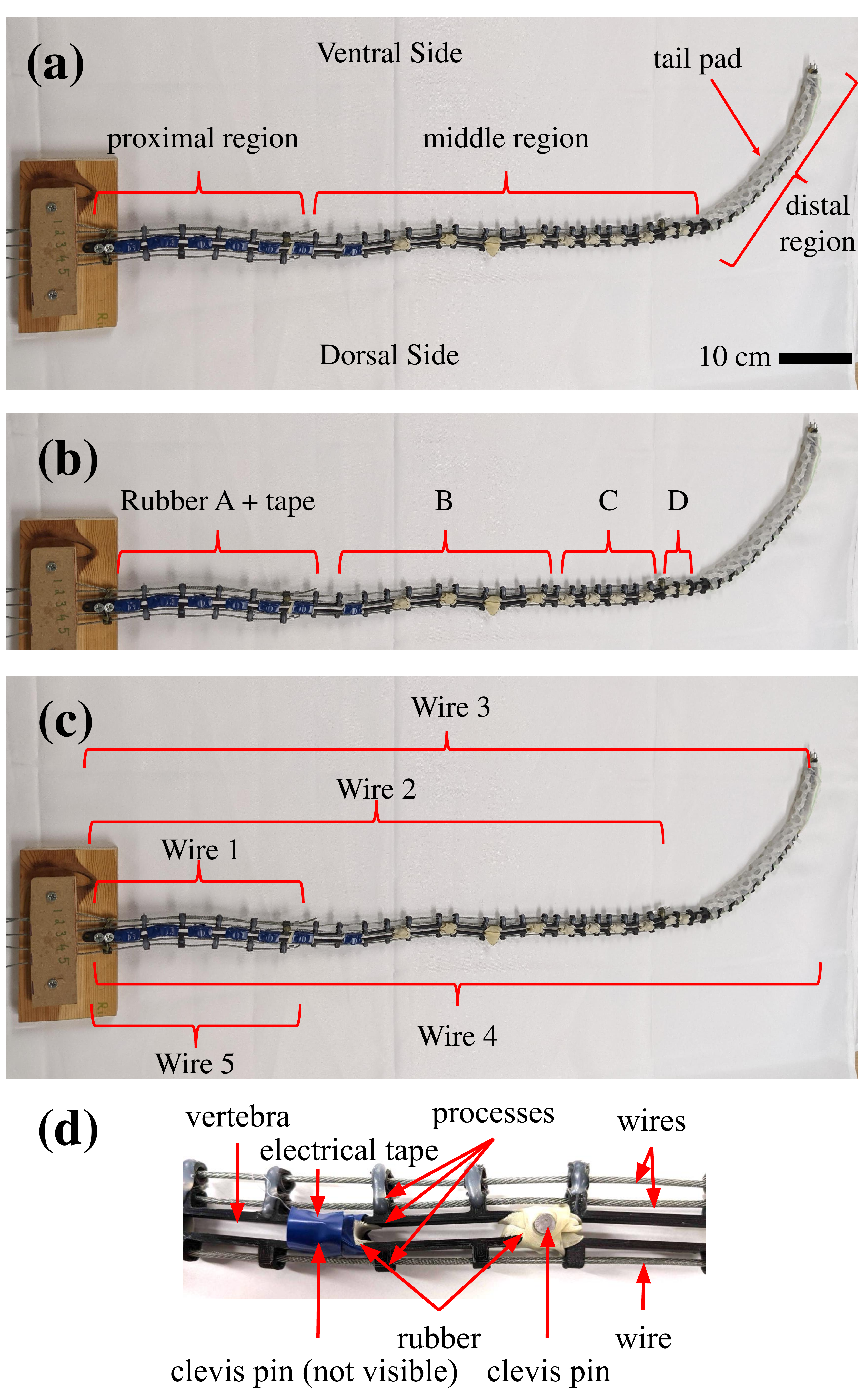}

  \caption{Design of our spider monkey tail-inspired robot. (a) The three major regions of the robot (proximal, middle, and distal), with the tail pad as part of the distal region. Similar to the biological spider monkey tail, grasping is designed to occur primarily on the ventral side (the side facing the front of the monkey). (b) Materials used to create decreasing stiffness along the length of the tail. Varying thicknesses A, B, C, and D of stretchable rubber (and electrical tape in the proximal region) are wrapped around each joint, with A as the thickest and D as the thinnest. (c) Identification of the five wires used to control movement of the robot. Wires 1 and 5 run the length of the proximal region, Wire 2 runs the length of the proximal and middle regions, and Wires 3 and 4 run the entire length of the tail. (d) At the transition between the proximal and middle regions, a detail view with identified elements: vertebrae with processes, pins, rubber, tape, and wires.
      }
      \label{Design}
      \vspace{-0.5cm}
  \end{figure}

In this section, we present the design of our tail robot (Figure~\ref{Design}). The robot structures around a vertebral column, with pin-jointed vertebrae that mimic a spider monkey's tail anatomy. Rubber and tape surrounding the pin joints generate varying degrees of stiffness along the length of the tail, and wire ropes enable manual operation of the tail. Finally, a skin-like silicone and fabric tail pad assists with grasping and manipulating objects.
  
\subsection{Vertebral Column}
Rigid, pin jointed vertebrae form the main structure of the tail robot. Each vertebra is manufactured from rigid polylactic acid plastic (Tough PLA, Ultimaker, Ultrecht, Netherlands) using a 3D printer (S5, Ultimaker). Lengths of the vertebrae vary along the length of the robot and approximately correspond to the lengths of the vertebrae in the biological spider monkey tail. Information about vertebra sizing was extrapolated from Chang and Ruch~\cite{Chang1947} as well as from CT scans of spider monkeys from MorphoSource's online database \cite{CTScan}. The three main regions of the tail, the tail pad, and the dorsal and ventral sides of the tail are indicated in Figure~\ref{Design}(a).

Each of the 29 3D printed vertebrae includes processes, or protrusions from the main column of the vertebra. These processes imitate the bony protrusions on biological vertebrae to which tendons attach. On the tail robot, control wires act like tendons and route through the processes.

In our robot, the proximal region is about 19 cm long and has 6 vertebrae, with wider processes than in the middle and distal regions. The middle region is about 42 cm long and has 10 vertebrae, with longer and slenderer vertebrae than in the proximal region. The distal region has 13 pin jointed vertebrae and 9 more loosely articulated vertebrae. The pin jointed section of the distal region is about 21 cm long, followed by a more flexible 11 cm section. The full tail length, excluding the base, is 93 cm.

The flexible 11 cm section of the distal region imitates the freely articulated bones of the flexible tip of the biological spider monkey tail. The last 9 vertebrae consist of 5 mm long tubes cut from 6.4 mm diameter plastic pneumatic tubing (UV-Resistant Soft PVC Plastic Tubing for Air and Water, 3/16" ID, 1/4" OD, McMaster-Carr Part 5231K35, Indiana, USA), through which the control wires route. To form each vertebra, the 5 mm sections of tubing are glued into pairs. A 12 cm length of soft PVC tubing (Clear Masterkleer Soft PVC Plastic Tubing for Air and Water, 1/32" ID, 3/32" OD, McMaster-Carr Part 5233K91, Indiana, USA) acts as a flexible backbone. The soft tubing connects with glue to the final 3D printed joint. 

\subsection{Nonuniform Stiffness Due to Soft Elements}
Rubber and electrical tape provide structural stiffness to enable control of the tail from its base. At the pin joints between individual links, a single layer of rubber wraps around the connection and affixes to the end of the metal clevis pin (1004-1045 Carbon Steel Clevis Pin, 3/16" Diameter, 5/16" Usable Length, McMaster-Carr Part 98306A100, Indiana, USA). In the base region, higher stiffness is created by wrapping a single layer of electrical tape around the rubber layer. The rubber decreases in thickness along the length of the tail so that structural stiffness decreases towards the tip.

Rubber of durometer 40A (medium soft) with the following thicknesses is used in the tail as indicated in Figure~\ref{Design}(b): A: thickness 0.76 mm; B: 0.51 mm; C: 0.36 mm; D: 0.30 mm (Super-Stretchable Abrasion-Resistant Natural Rubber Sample Packs, McMaster-Carr Part 8611K222, Indiana, USA). Since the rubber fits tightly against the vertebral column---especially due to the added electrical tape in the proximal section---it does not interfere with the control wires.

\subsection{Force Transmission and Actuation Mechanism}

Five nylon-coated 7x7 strand galvanized steel wire ropes (Coated Wire Rope - Not for Lifting, 0.047" Diameter, 1/16" Diameter with Coating, Lubricated, McMaster-Carr Part 8923T115, Indiana, USA), cuffed at their ends and routed through the vertebral processes, can move different sections of the tail. Two antagonistic pairs of short and long control wires exist on the dorsal and ventral sides of the tail. An additional mid-length wire exists on the ventral side to provide additional control, since more finely controlled tail articulation was observed on the ventral tail sides of the spider monkeys at the Potawatomi Zoo. Because the structural stiffness of the tail diminishes towards the tip, a single wire running the full length of the tail can control the movement of the tip of the tail without moving the base. Figure~\ref{Design}(c) details the wire placement on the tail.

\subsection{Tail Pad}

In its biological form, the spider monkey tail's distal region includes a tail pad, a hairless portion on the ventral side that aids in gripping. With no tail pad in place on the robot, the robot provided negligible resistance to the removal of an object from its grasp. A tail pad was fabricated from molded silicone of durometer 30 (Ecoflex 00-30, Smooth-On, Pennsylvania, USA) fixed to a ripstop nylon fabric sleeve (1.1 oz MTN Silnylon 6.6, Ripstop by the Roll, North Carolina, USA) using Sil-Poxy adhesive. Since the smooth side of the ripstop nylon contacts the wires, the added tail pad does not inhibit movement of the tail's control wires. 

\section{Experiments and Results}
To gauge the ability of the tail robot to perform bio-inspired tasks, we evaluated its mobility as well as its ability to manipulate and grasp objects of various sizes and shapes.

\subsection{Tail Mobility}

\begin{figure}[tb]
      \centering
      \includegraphics[width=0.82\columnwidth]{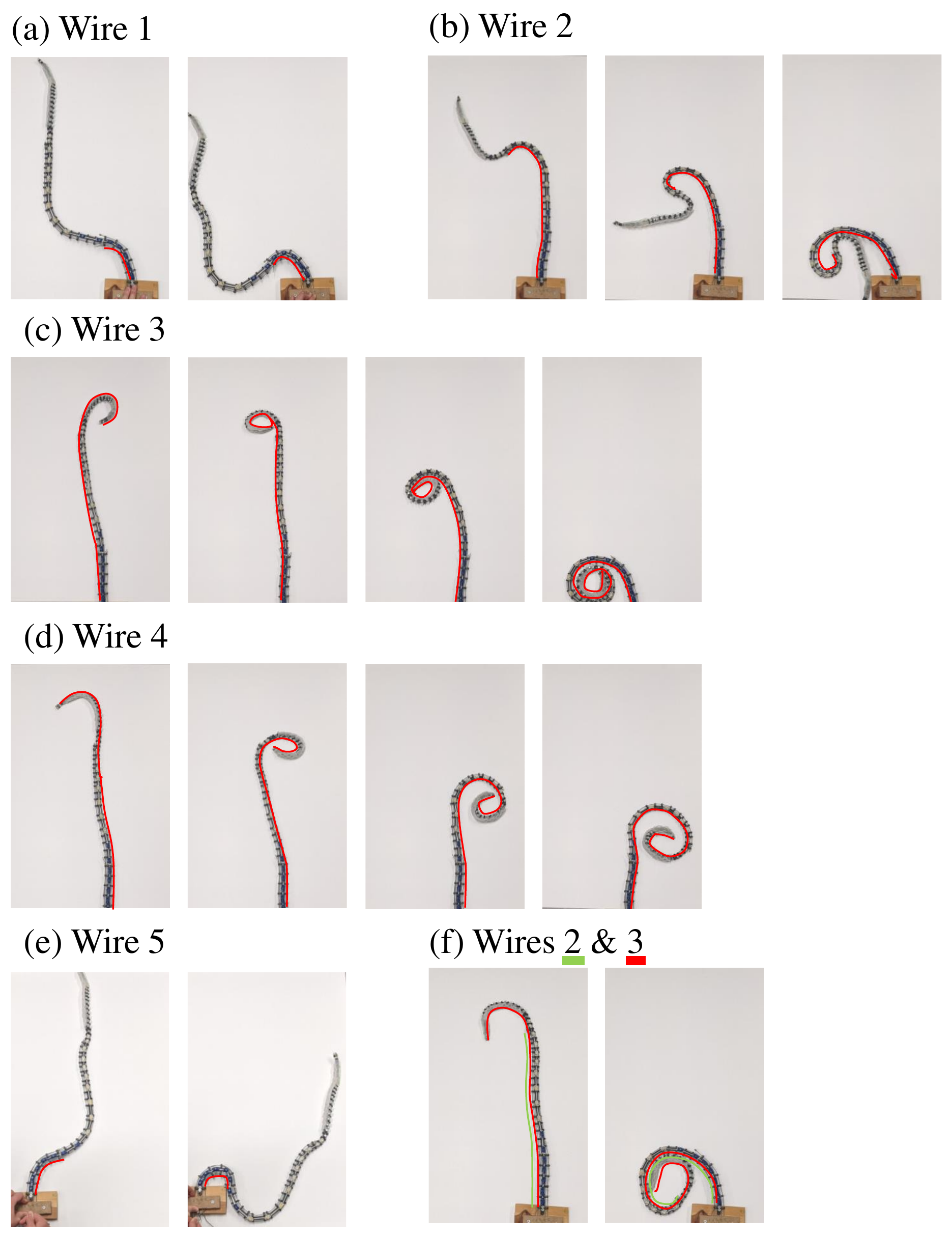}
      \caption{Planar robot motions corresponding to actuation of various individually-operated control wires. (a) Wire 1 moves the proximal region towards the ventral (left) side. (b) Wire 2 first moves the middle region and then the proximal region towards the ventral side. (c) Wire 3 first moves the distal region and then the middle and proximal regions towards the ventral side. (d) Wire 4 first moves the distal region and then the middle and proximal regions towards the dorsal (right) side. (e) Wire 5 moves the proximal region towards the dorsal side. (f) Wires can be actuated in combination. For instance, Wires 2 and 3 can together curl the robot towards the ventral side, resulting in a different configuration than with the movement of Wire 3 alone. For all, gravity is perpendicular to the plane of motion. 
      }
      \label{Actuation}
      \vspace{-0.5cm}
  \end{figure}
  
Figure~\ref{Actuation} demonstrates how pulling on different wires corresponds to movement in the tail. 
For the two longest wires (Wires 3 and 4), pulling (or pushing) the wire can curl the tip of the tail in clockwise or counterclockwise directions. Pushing is most easily applied to these wires due to the lower stiffness at the robot tip. Two antagonistic pairs of wires (Wires 1 and 5; Wires 3 and 4) generate mirrored movements, with Wires 1 and 5 increasing the workspace of the tail tip by curving the proximal region of the tail. The mid-length Wire 2 provides a greater range of movement on the ventral side of the tail.
Movement of the wires in combination with each other can produce a variety of shapes, as demonstrated with Wires 2 and 3 combined, which causes the tail to form a larger loop than actuation of Wire 3 alone.

\subsection{Grasping Force vs. Object Diameter}

Stuart et al.~\cite{Stuart2017} note that resisting pull-out force gives a useful measurement of a robot's capability to handle objects. 

\begin{figure}[tb]
      \centering
      \includegraphics[width=\columnwidth]{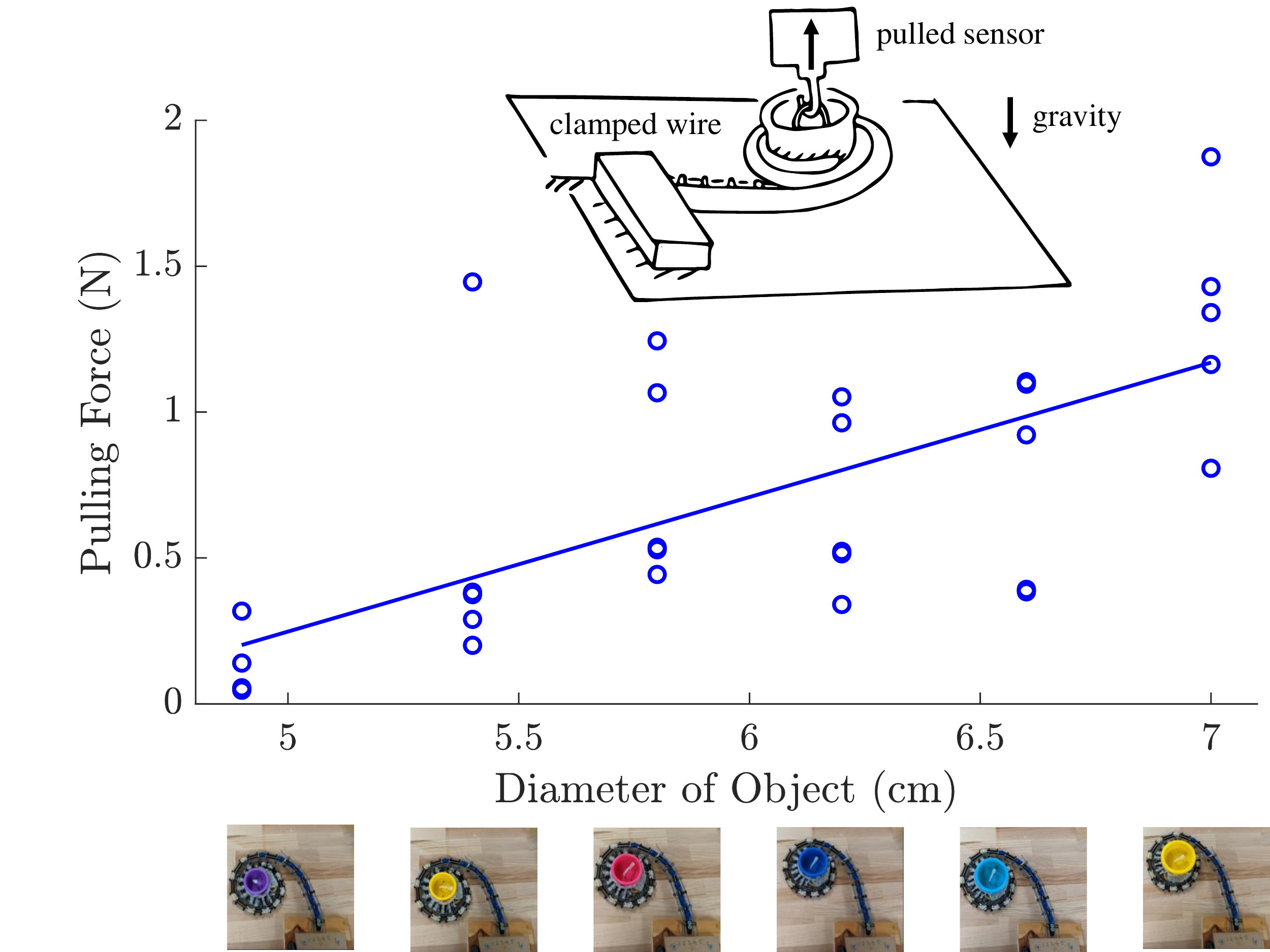}
     
      \caption{Pulling force required to remove circular objects from the robot's grasp vs. diameter of the object. Object weights were subtracted from raw sensor data to create the plot. Data is shown for five trials at each diameter (circles), alongside a best fit line for the data (solid line). The robot is shown grasping each object below its corresponding data points. The force is roughly linear with diameter, which indicates that, as the contact area between the robot and the object increases, the force increases. The experimental setup is illustrated above the plot. 
      }
      \label{ForceDiameter}
      \vspace{-0.5cm}
  \end{figure}

To determine the force that the robot could exert in grasping, a set of 6 round toy cups of varying diameters (4.9, 5.4, 5.8, 6.2, 6.6, and 7.0 cm) and corresponding masses (13.0, 15.8, 18.3, 19.7, 23.9, and 26.5 g) was used in conjunction with a MARK10 M3-20 force gauge sensor, illustrated in Figure~\ref{ForceDiameter}. 

Five trials were performed for each of the 6 cups. In each trial, Wire 3 of the robot was pulled to fully grasp the object, the wire was clamped in place, and then the object was pulled from the grasp of the tail using the sensor's hook attachment. Peak pulling force from each trial was collected five times per object. 

Pulling force with gravitational force acting on the cup subtracted is plotted against corresponding object diameter in Figure~\ref{ForceDiameter}. The solid line on the plot shows the best linear fit over the measured data points. The equation of this estimated relationship between the peak pulling force ($F$) and the object diameter ($d$) is
\begin{equation} \label{eq:1}
    F = 0.461 d -2.057 .
\end{equation}

The results in Figure~\ref{ForceDiameter} demonstrate a correlation between increased diameter of the object and increased force required to pull the object from the tail's grasp. The data suggests that for the tail, grasping force is largely a function of friction corresponding to the surface area contacted between the silicone tail pad and the grasped object.

\subsection{Grasping Force vs. Object Shape}

To determine the force that the robot could exert in grasping objects of varying shapes, a set of 4 geometric shapes was created: an equilateral triangle, square, and pentagon inscribed into a 9 cm diameter circle, as well as the 9 cm circle. Following the same procedure as illustrated in Figure~\ref{ForceDiameter}, the force gauge sensor was used to measure the force to remove an object from the grasp of the tail. The weight from each object's mass (22.6, 35.1, 40.4, and 55.2 g) was subtracted to determine pulling force.

\begin{figure}[tb]
      \centering
     \includegraphics[width=\columnwidth]{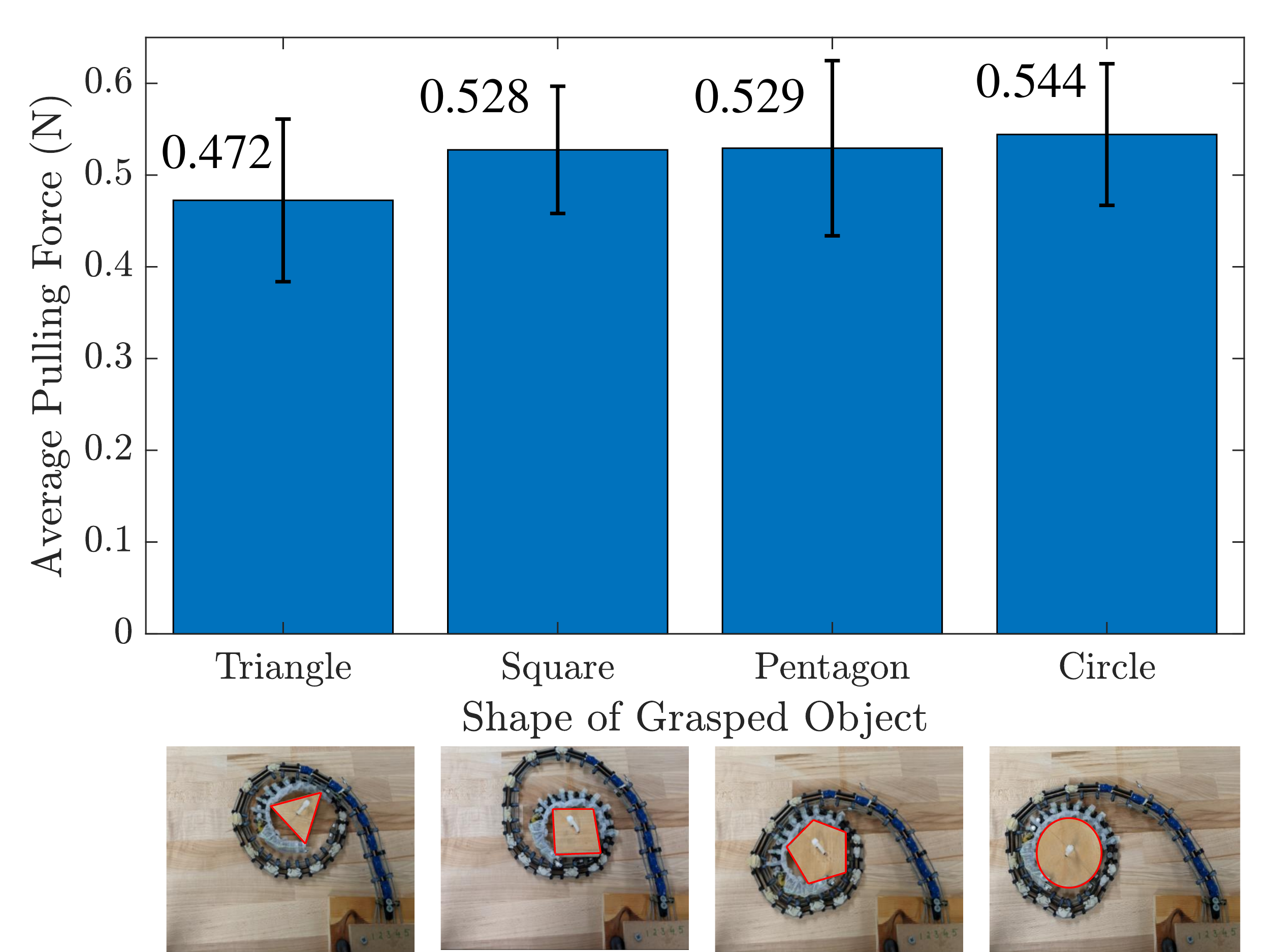}
     
      \caption{Average pulling force required to remove an object from the robot's grasp vs. shape of the object (triangle, square, pentagon, and circle). Object weights were subtracted from raw sensor data to create the plot. Data is shown for five trials for each shape as a blue bar for the average value and black bars representing standard error. Below the plot, the robot is shown grasping each object with the outline of the object highlighted in red. The force generally increases with the number of sides of the object, but it is not a simple function of contact area.
      }
      \label{ForceShape}
      \vspace{-0.5cm}
  \end{figure}

Average peak pulling force from five trials per object was plotted against the corresponding shape in Figure~\ref{ForceShape}. Based on these results, a general, though seemingly nonlinear, correlation between grasping force and degree of contact between the tail and the grasped object can be seen, with force increasing alongside the number of edges of the geometric shape. As before, these results suggest that the grasping force of the tail is generally a function of the contact area between the silicone tail pad and the object; further evaluation could explore the impact of the deformation of the tail pad around objects with corners of different angles, for instance.

\section{Demonstration}

We demonstrate the capability of the tail robot to manipulate and grasp objects of various sizes and to navigate around objects in imitation of the spider monkey. We also demonstrate retrieval of an object through a low passageway.

\subsection{Grasping Objects of Different Sizes}

Grasping, manipulation, and release of a variety of supports are key behaviors of the spider monkey tail in nature. As the spider monkey navigates its tree canopy habitat, it takes advantage of a variety of supports, ranging from narrow twigs and vines to larger tree limbs. As a bio-inspired design, the tail-inspired robot seeks to imitate this capability.

\begin{figure}[tb]
      \centering
      \includegraphics[width=0.95\columnwidth]{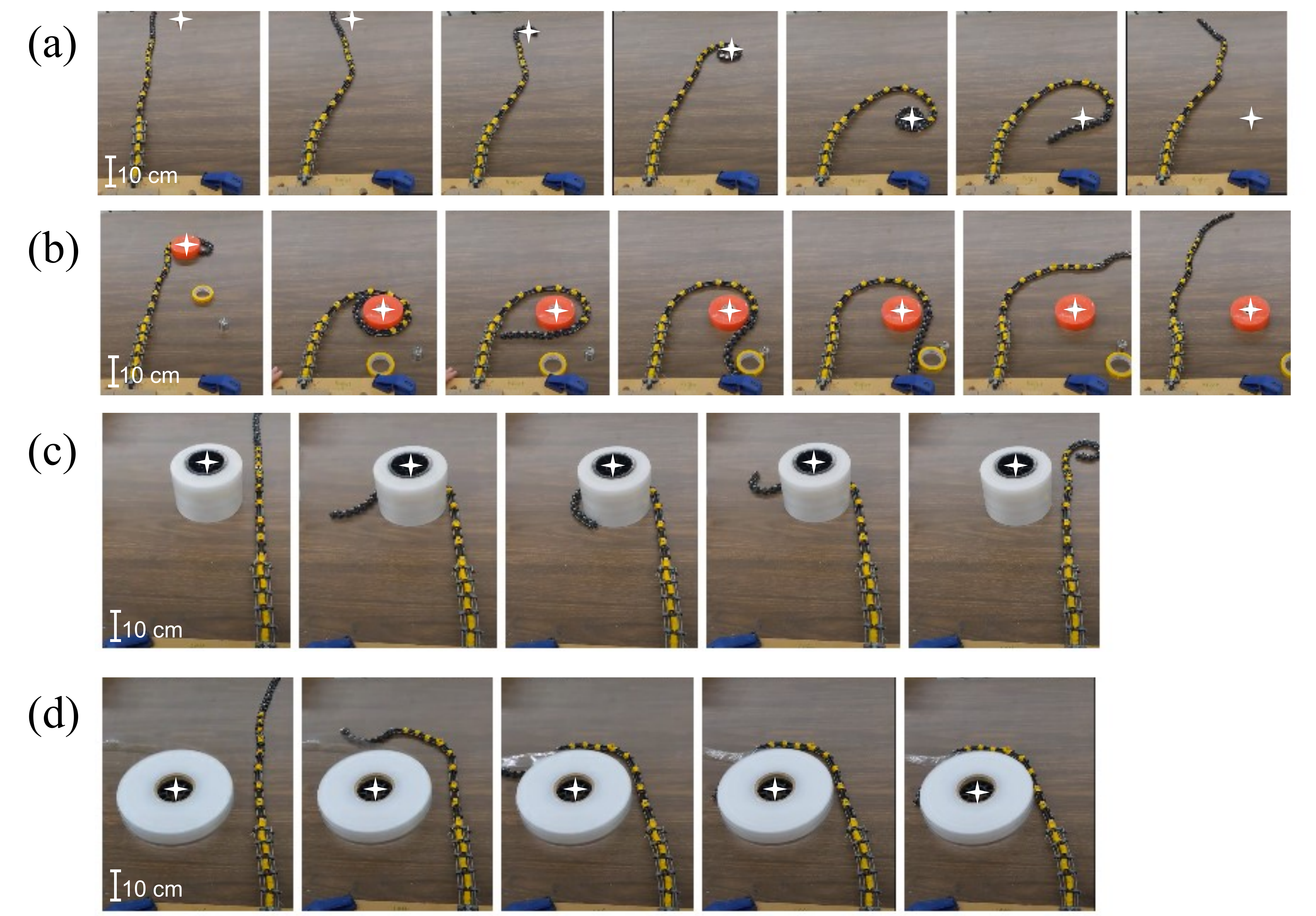}
     
      \caption{Demonstration of grasping, moving, and releasing objects of various diameters, with a star indicating each object's center. (a) The robot grasps, moves, and releases a 2.0 cm diameter object. (b) The robot grasps, moves, and releases a 9.3 cm object. (c) The robot grasps and releases a 15.0 cm diameter object. (d) The robot grasps and releases a 23.5 cm diameter object.
      }
      \label{DemoManipulation}
      \vspace{-0.5cm}
  \end{figure}

An earlier iteration of the tail robot (which lacked a tail pad but otherwise functionally mirrored the existing robot) demonstrated grasping and manipulation of objects via the control inputs of pushing and pulling on the wires. Figure~\ref{DemoManipulation} demonstrates manipulation around objects of various sizes. In Figure~\ref{DemoManipulation}(a), the tail robot encloses, manipulates, and then releases a 2.0 cm diameter object. In Figure~\ref{DemoManipulation}(b), the tail grasps a 9.3 cm diameter object, drags it across a surface towards the base of the tail, then releases the grasp while avoiding nearby obstacles. In Figure~\ref{DemoManipulation}(c), the tail secures itself around a 15.0 cm diameter object. The distal region of the tail freely articulates apart from the rest of the tail to ensure a secure position. Next, the tail releases the object. Figure~\ref{DemoManipulation}(d) demonstrates motions similar to the prior instance, but with a 23.5 cm diameter object.

\subsection{Retrieving Object from Under a Door}
\begin{figure}[tb]
      \centering
      \includegraphics[width=0.6\columnwidth]{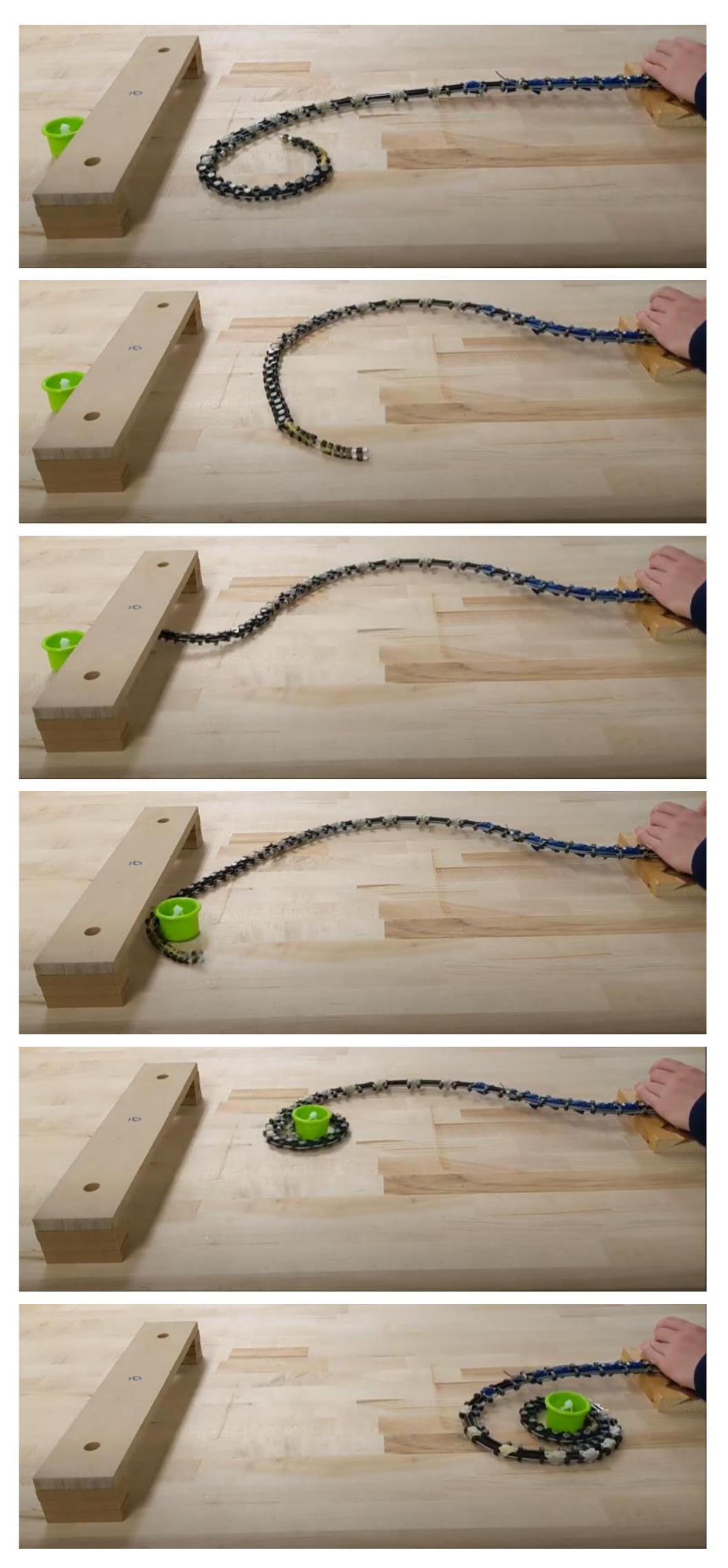}
     
      \caption{Demonstration of retrieval of an object after passing under a low passageway. Operated from the base via wires, the tail robot successfully reaches under a low passageway to retrieve a plastic cup and bring it back towards the base. This task could be challenging for a biological monkey's arm or a bulkier robot arm but is straightforward for the tail robot due to its low profile and versatile grasping ability.
      }
      \label{DemoPassageway}
      \vspace{-0.5cm}
  \end{figure}

Kendall and Nelson~\cite{Nelson2018} documented the ability of spider monkey tails to retrieve objects only reachable by the tail, not reachable by the fore- or hindlimbs. To demonstrate similar retrieval with the tail-inspired robot, the base of the robot was fixed in place and a small object (a plastic toy cup) was placed on the opposite side of a low passageway. Operating the tail via wires, the tail reached under the passageway, grasped the cup, and pulled it out through the passageway, towards the tail's base. Figure~\ref{DemoPassageway} documents the series of movements in this retrieval demonstration, mimicking the desired function the biological tail performs.

\section{Conclusion and Future Work}

We presented a tail robot that imitates the morphology of the spider monkey tail. Nonuniform stiffness along the length of the tail, created via rubber and tape, makes possible a slim and lightweight tail design. Five wires, manually operated in this iteration, represent a relatively simple control scheme. In the plane, the tail robot demonstrates abilities similar to those of the spider monkey: it grasps and releases objects of varying diameters and shapes. The robot also navigates around obstacles and through low passageways. In the robot, grasping force correlates to the friction created by contact area between the grasped object and the bio-inspired tail pad.

Since the robot is currently constrained to planar movement, future research will focus on making the tail robot operable in 3D space. We also plan to add motors to the tail rather than operating wires manually. Future observations with the Potawatomi Zoo will inform the ongoing development of the morphology of the tail pad. We will also explore how the tail robot might interface with a larger robotic system---for instance, with a quadruped robot---to assist with movement throughout a complex terrain. Our work will seek to make a robotic prehensile tail feasible and useful in a variety of environments.

\section{Acknowledgements}
We would like to thank the Potawatomi Zoo (South Bend, IN, USA) for their generous support of this project and for welcoming us to observe the resident spider monkeys.


\printbibliography

@INPROCEEDINGS{Arachchige2022,
  author = {D. D. K. Arachchige and I. S. Godage},
  title = {Hybrid Soft Robots Incorporating Soft and Stiff Elements},
  BOOKTITLE = {IEEE International Conference on Soft Robotics},
  YEAR = { 2022 },
  pages = {267-272}
}

@article{Mochiyama2022,
  title={Ostrich-Inspired Soft Robotics: A Flexible Bipedal Manipulator for Aggressive Physical Interaction},
  author = {H. Mochiyama and M. Gunji and R. Niiyama},
  journal={Journal of Robotics and Mechatronics},
  year={2022},
  volume = {34},
  pages = {212-218},
  number = {2},
}

@ARTICLE{Holt2017,
  author = {J. D. Holt},
  title = {Design and Testing of a Biomimetic Pneumatic Actuated Seahorse Tail Inspired Robot},
  note={M.S. thesis, Clemson University, 2017.}

}

@ARTICLE{Matsuda2022,
  author = {R. Matsuda and U. K. Mavinkurve and A. Kanada and K. Honda and Y. Nakashima and M. Yamamoto},
  title = {A Woodpecker's Tongue-inspired, Bendable and Extendable Robot Manipulator with Structural Stiffness},
  journal = {IEEE Robotics and Automation Letters},
  year = {2022},
  volume = {7},
  pages = {3334-3341},
  number = {2},
}

@ARTICLE{Stuart2017,
  author = {H. Stuart and S. Wang and O. Khatib and M. R. Cutkosky},
  title = {The Ocean One hands: An adaptive design for robust marine manipulation},
  journal = {International Journal of Robotics Research},
  year = {2017},
  volume = {36},
  pages = {150-166},
  number = {2},
}

@ARTICLE{Chang1947,
  author = {H. T. Chang and T. C. Rush},
  title = {Morphology of the Spinal Cord, Spinal Nerves, Caudal Plexus, Tail Segmentation, and Caudal Musculature of the Spider Monkey},
  journal = {Yale Journal of Biology and Medicine},
  year = {1947},
  volume = {19},
  pages = {345-377},
  number = {3},
}

@ARTICLE{Nelson2018,
  author = {E. L. Nelson and G. A. Jendall and D. M. Fragaszy},
  title = {Goal-Directed Tail Use in {C}olombian Spider Monkeys ({A}teles fusciceps rufiventris) Is Highly Lateralized},
  journal = {Journal of Comparative Psychology},
  year = {2018},
  volume = {132},
  pages = {40-47},
  number = {1},
}

@INPROCEEDINGS{Hannan2000,
  author = {M. W. Hannan and I. D. Walker},
  title = {Analysis and Initial Experiments for a Novel Elephant's Trunk Robot},
  BOOKTITLE = {IEEE/RSJ International Conference on Intelligent Robots and Systems},
  YEAR = { 2000 },
  pages = {330-337}
}

@INPROCEEDINGS{Rosenberger2008,
  author = {A. L. Rosenberger and L. Halenar and S. B. Cooke and W. C. Hartwig},
  title = {Morphology and evolution of the spider monkey, genus {A}teles},
  note={in \textit{Spider Monkeys: Behavior, Ecology and Evolution of the Genus Ateles}, 2008, pp. 19-49}
}

@INPROCEEDINGS{Youlatos2008,
  author = {D. Youlatos},
  title = {Locomotion and positional behavior of spider monkeys},
  note={in \textit{Spider Monkeys: Behavior, Ecology and Evolution of the Genus Ateles}, 2008, pp. 185-219}
}

@ARTICLE{CTScan,
  author = {MorphoSource},
  title = {Whole Body \textit{Ateles belzebuth}},
  BOOKTITLE = {MorphoSource},
  YEAR = { 2000 },
  note={[Online]. Available: https://www.idigbio.org/portal/records/10812957-753c-4b3f-ae3e-947ef01d799f.}
}

@ARTICLE{FinalMonkeyPic,
  author = {S. J. Tonge},
  title = {Image of Brown Spider Monkey},
  YEAR = { 2014 },
  note={[Online]. Available: https://eol.org/media/8897642.}
}


\end{document}